%% file: main.tex
\documentclass{src/bmvc2k}

\title{A Better Use of Audio-Visual Cues: \\Dense Video Captioning with Bi-modal Transformer}

\addauthor{Vladimir Iashin}{vladimir.iashin@tuni.fi}{1}
\addauthor{Esa Rahtu}{esa.rahtu@tuni.fi}{1}

\addinstitution{
 Computing Sciences\\
 Tampere University\\
 Tampere, Finland
}

\runninghead{Iashin, Rahtu}{A Better Use of Audio-Visual Cues}

\def\eg{\emph{e.g}\bmvaOneDot}

\def\etal{\emph{et al}\bmvaOneDot}

\newcommand{\comma}{, \, }
\newcommand{\R}{\mathrm{R}}
\newcommand{\inR}{\in \mathrm{R}}
\usepackage{booktabs}
\usepackage{multirow}

\begin{document}

\maketitle

\begin{abstract}
Dense video captioning aims to localize and describe important events in untrimmed videos. Existing methods mainly tackle this task by exploiting only visual features, while completely neglecting the audio track. Only a few prior works have utilized both modalities, yet they show poor results or demonstrate the importance on a dataset with a specific domain. In this paper, we introduce \textit{Bi-modal Transformer} which generalizes the Transformer architecture for a bi-modal input. We show the effectiveness of the proposed model with audio and visual modalities on the dense video captioning task, yet the module is capable of digesting any two modalities in a sequence-to-sequence task. We also show that the pre-trained bi-modal encoder as a part of the bi-modal transformer can be used as a feature extractor for a simple proposal generation module. The performance is demonstrated on a challenging \textit{ActivityNet Captions} dataset where our model achieves outstanding performance. The code is available: \textbf{\texttt{\color{blue} v-iashin.github.io/bmt}}
\end{abstract}

\section{Introduction \label{sec:intro}}
Current video sharing platforms contain a large amount of video material. The ability to generate descriptions of this content would be highly valuable for many tasks, such as content-based retrieval or recommendation \cite{Tran2015,HowTo100M_Miech2019}. Moreover, they would enable visually-impaired people to consume video material and improve their quality of life \cite{LSMDC_Rohrbach_2017}.

\begin{figure}
\begin{center}
\includegraphics[width=\textwidth]{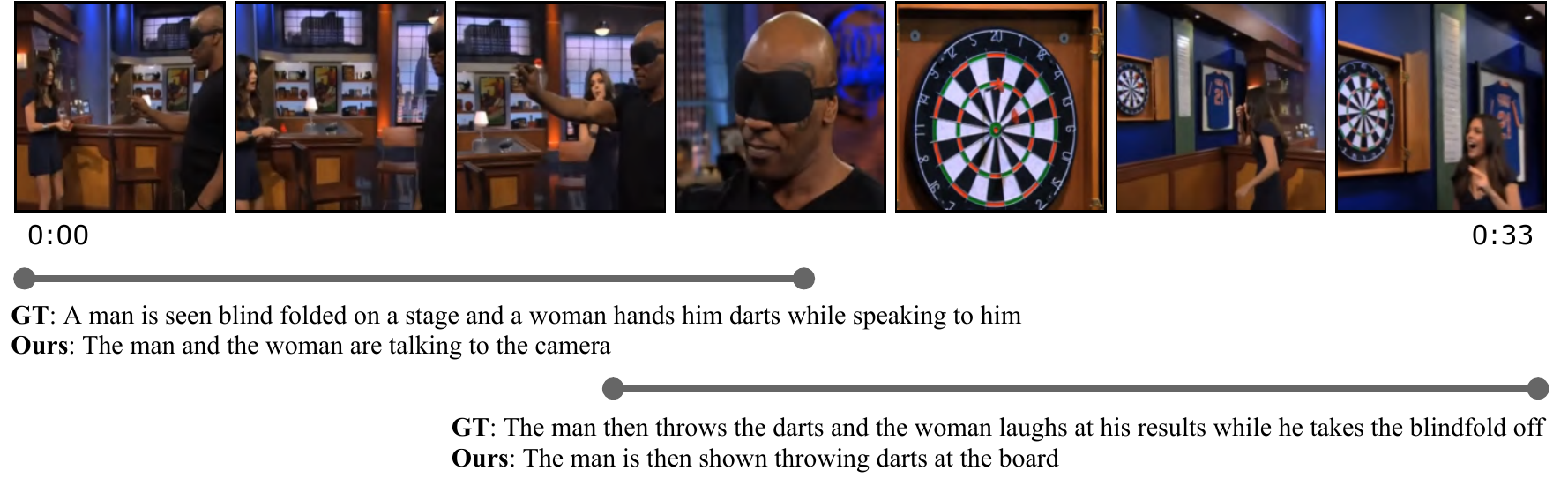}
\end{center}
\vspace{-1em}
\caption{Example video with the predictions of our model alongside the ground truth. \label{fig:teaser}}
\end{figure}

This kind of video descriptions are usually provided as natural language sentences or \textit{captions}, a compact and intuitive format and, most importantly, can be digested by humans. Early works \cite{Yao2015,Video2Text_Venugopalan2015,Venugopalan2015b,Yu2016} described the video content with only one sentence, which might be too ``sparse'' for long videos -- one might try to think up a relatively short sentence which describes the whole film. To mitigate this issue, \cite{Krishna2017} proposed \textit{dense video captioning} which requires a model to, first, localize ``events'', and, then, to produce one-sentence description for each of them instead of generating one caption for the entire film (see Fig.~\ref{fig:teaser}).

The task is usually formulated as a \textit{sequence-to-sequence} (video to caption) task. Therefore, the progress in the field is significantly influenced by advances in machine translation. Hence, many models rely on an encoder-decoder architecture which consists of two \textit{recurrent neural networks} (RNNs) or, recently-proposed \textit{Transformer}-like model \cite{transformer_Vaswani2017}. An event localization module usually utilizes an RNN structure which first encodes the input to produce a hidden representation and, then, makes predictions using this representation.

Considering the natural co-occurrence of visual and audio tracks in a video and the fact that human perception is multi-modal, recent advances in deep learning practice audio-visual training \cite{AV_Scene_Owens_2018,TheSoundPixels_Zhao_2018,AttentionClusters_Long_2018,VisGrounded_Zhu_2020,Separating_Zhu_2020}. Yet, most of the existing works on dense video captioning employ only visual inputs. In this work, we address this issue by introducing a novel bi-modal transformer with the multi-headed proposal generator. Our captioning module is inspired by the transformer architecture and, more precisely, how the attention module fuses the information from both sequences. While an efficient object detector \textit{YOLO} \cite{YOLOv3_Redmon_2018} inspires the design of each proposal head in the bi-modal multi-headed proposal generator.

The proposed method effectively utilizes audio and visual cues. We demonstrate the performance of our model on the challenging open-domain ActivityNet Captions dataset \cite{Krishna2017}. The results show the state-of-the-art performance of our bi-modal dense video captioning module as well as our bi-modal proposal generator on BLEU@3--4 and F1 metrics.

\section{Related Work}

The dense video captioning task requires a model to, first, localize events within a video and, then, to produce a textual one-sentence description of what is happening during the event. The dense video captioning task branches out from the \textit{video captioning} which task is to caption a video without localizing the event. The video captioning field evolved from hand-crafted rule models \cite{Kojima2002,MunWaiLee2008,Das2013} to \textit{encoder-decoder} architectures \cite{Yao2015,Video2Text_Venugopalan2015,Venugopalan2015b,Yu2016} inspired by advances in machine translation \cite{Rohrbach2013}. Later, the captioning models were further enhanced by \textit{semantic tagging} \cite{Gan2017,Pan2017}, \textit{reinforcement learning} \cite{Wang2018}, \textit{attention} \cite{Yan2019}, \textit{extended memory} \cite{Wang2018a,Pei2019}, and other modalities \cite{Xu2017,Hori2017,Wang2018b,Hao2017}.

\subsection{Dense Video Captioning}

The task of dense video captioning, as well as a test-bed, ActivityNet Captions dataset, were introduced by Krishna \etal~\cite{Krishna2017} who utilized the idea of the \textit{Deep Action Proposals} network \cite{daps_Escorcia2016} to generate event proposals and an LSTM network to encode the context and generate captions. The idea of context-awareness was further developed in \cite{bafcg_Wang2018n} who employed a bi-directional variant of \textit{Single-Stream Temporal} Action proposal network (SST) \cite{SST_Buch2017} which makes better use of the video context, an LSTM network with \textit{attentive fusion and context gating} was used to generate context-aware captions. Zhou \etal~\cite{masked_transformer_Zhou2018} adapted \textit{Transformer} architecture \cite{transformer_Vaswani2017} to tackle the task and used transformer \textit{encoder}'s output as input to a modification of \textit{ProcNets} \cite{ProcNets_2018_Zhou} to generate proposals.

Recently, the idea of reinforcement learning was found to be beneficial for image captioning (\textit{Self-critical Sequence Training} (SCST)) \cite{SCST_Rennie_2017} and, hence, applied in dense video captioning as well. More precisely, the SCST was used in a captioning module to optimize the non-differentiable target metric, \eg METEOR \cite{METEOR_Denkowski2014}. Specifically, Li \etal~\cite{Li2018} integrated the reward system and enriched \textit{Single-Shot-Detector}-like structure \cite{SSD_Liu2016} with descriptiveness regression for proposal generation. Similarly, Xiong \etal~\cite{MFT_Xiong2018} used an LSTM network trained with the sentence- and paragraph-level rewards for maintaining coherent and concise story-telling, while the event proposal module was adopted from \textit{Structured Segment Networks} \cite{Zhao2017}. Mun \etal~\cite{Streamlined_Mun2019} further developed the idea of coherent captioning by observing the overall context and optimizing two-level rewards, an SST module is used for proposal generation, and a \textit{Pointer Network} \cite{PointerNet_Vinyals_2015} to distill proposal candidates.

Another direction of research relies on weak supervision which is designed to mitigate the problem of laborious annotation of the datasets. To this end, Duan \etal~\cite{WeaklySupervDVC_Duan_2018} proposed an \textit{autoencoder} architecture which generates proposals and, then, captions them while being supervised only with a set of non-localized captions in a \textit{cycle-consistency} manner. However, the results appeared to be far from the supervised methods.

\subsection{Multi-modal Dense Video Captioning}

It is natural to assume that, besides visual information, a video understanding system might benefit from the cues contained in other modalities like audio \cite{Rahman2019}, speech (subtitles) \cite{DenseProcCap_Shi_2019}, or both \cite{mdvc_Iashin_2020}. Specifically, Rahman \etal~\cite{Rahman2019} were the first to include audio modality into the dense video captioning set up. They borrowed the idea of cycle-consistency from \cite{WeaklySupervDVC_Duan_2018} and employed \textit{multi-modal Tucker decomposition} \cite{MUTAN_Hedi_2017} to combine information from both modalities and pass it to a \textit{GRU}-based \cite{GRU_Chung_2014} caption decoder. However, since the model is trained in a weakly supervised setting, the results do not reach the performance of the supervised~models.

Shi \etal~\cite{DenseProcCap_Shi_2019} proposed to utilize the corresponding speech along with frame features to further improve captioning performance on cooking videos. They suggested  employing a transformer's encoder to encode video frames and subtitle \textit{embeddings} produced by a pre-trained \textit{BERT} model \cite{BERT_Devlin_2018}. Next, an LSTM generates proposals, and the other two LSTMs were used for the encoder-decoder captioning module. Despite the significant gains in captioning performance, we believe these findings are not conclusive as instructional videos is an ill-suited domain to show the benefits of the speech modality for a captioning task since subtitles alone can be a very accurate proxy for captions in such videos (see \cite{HowTo100M_Miech2019}).

In contrast, Iashin \etal~\cite{mdvc_Iashin_2020} showed the importance of the speech modality on a free-domain dataset. They proposed to train three transformers for each modality individually and fuse features by concatenation before predicting the next caption word while borrowing the proposal generator from \cite{bafcg_Wang2018n}. However, the suggested approach for feature fusion is rather straightforward and inefficient. Moreover, the adopted proposal generator is based solely on video features which contrasts with the idea of the dense video captioning task.

Our method is mostly similar to \cite{mdvc_Iashin_2020}, yet we show significantly better results on the task while utilizing only visual and audio cues. Besides, our proposal generator does employ both modalities and significantly outperforms the state-of-the-art. Furthermore, we present a single model which utilizes bi-modal encoder for both: the proposal generator and captioning module, making it an elegant approach for the dense video captioning task.

\begin{figure}
\begin{center}
\includegraphics[width=0.96\textwidth]{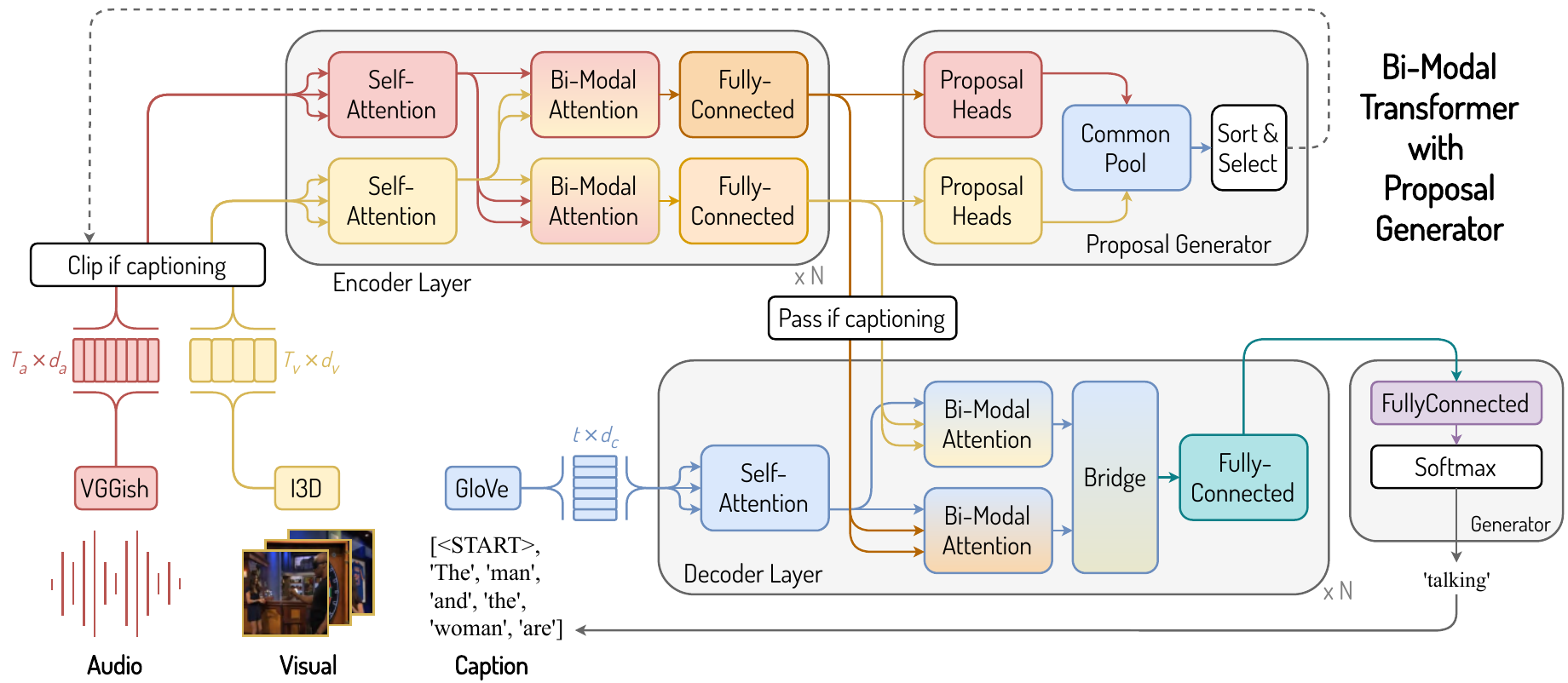}
\end{center}
\vspace{-1.5em}
\caption{\small The design of Bi-modal Transformer with Multi-headed Proposal Generator. The proposed model inputs features extracted by VGGish, I3D, and GloVe pre-trained models (bottom left). Then, the bi-modal encoder with N layers processes the audio and visual features and passes its bi-modal representation to the proposal generator (top). After, the generated proposals are used to clip the input features (left). The clipped features are passed through the encoder again. The output of the encoder, then, is used at every layer (N) of the bi-modal decoder (bottom). The decoder attends to the bi-modal encoder's representation as well as the previous caption words and produces its internal representation of the context. This representation is passed to the generator (right) to generate the next word. Residual connections are removed for clarity. Best viewed in color. \label{fig:bi_modal_transformer}}
\end{figure}


\section{Our Framework}

Our approach consists of two parts: the \textit{bi-modal transformer} and \textit{multi-headed proposal generator} (see Fig.~\ref{fig:bi_modal_transformer}). The model expects the input to be a set of continuous features stacked together in a sequence. To represent a visual stream, we use a pre-trained \textit{Inflated 3D} (I3D) network \cite{i3d_Carreira_2017} while for the audio stream we employ pre-trained \textit{VGGish} \cite{Hershey2017}, the tokens (roughly, words) are embedded with pre-trained \textit{GloVe} \cite{Glove_Pennington_2014} (see Sec.~\ref{sec:feature_extraction} for implementation details). Also, since the transformer is \textit{permutation invariant} it has no sense of recurrence. Thus, the order of features within a sequence is preserved by adding the \textit{positional encoding} to the output of the embedding layers. Following \cite{transformer_Vaswani2017}, we use \textit{cosine} and \textit{sine} functions.

Next, the audio and visual sequences, are passed through the transformer's bi-modal \textit{N}-layered \textit{encoder} to produce bi-modal sequence representations utilizing novel \textit{bi-modal multi-headed attention} blocks to fuse the features from both sequences. Then, the novel proposal generator utilizes these features to generate proposals and their confidence scores. After, a pre-defined number of most confident proposals are selected to clip the input feature sequences. Next, the clipped features are processed with the encoder to re-represent the features considering  only the features which are left after clipping.

The bi-modal encoder's representation is used at every layer in the bi-modal \textit{decoder}. Concretely, the encoder's outputs are passed to the corresponding bi-modal attention blocks in the decoder layer along with the representation of the previously generated caption words. The last-layer representation of the decoder is used in the \textit{generator} where the next caption word is produced. To avoid an empty input to the decoder in the beginning, a special \textit{start-token} is used. The caption is generated word-by-word until a special \textit{end-token} is sampled.

This section, first, presents the design of the captioning module (Sec.~\ref{sec:captioning_module}) and, second, the proposal generator (Sec.~\ref{sec:prop_generator}) while the training procedure is explained in Sec.~\ref{sec:train_proc}.

\subsection{Captioning Module\label{sec:captioning_module}}

The task of dense video captioning requires to produce a caption for each proposal. Therefore, \textit{bi-modal encoder} inputs audio $A$ and visual $V$ feature sequences which temporally correspond to the proposal and outputs two sequences: audio-attended visual features $V^a$ and visual-attended audio features $A^v$. These features are used by the \textit{bi-modal decoder} which attends to these features and the previous caption words $(c_1 \comma c_2 \comma \dots \comma c_t)$. Finally, the bi-modal decoder outputs the representation which is employed to model a distribution of the next caption word $(c_{t+1})$ over the vocabulary. The proposal index is omitted for clarity.

\vspace{-1.4ex}\paragraph{Bi-modal Encoder \label{sec:encoder}}
In contrast to the encoder in \cite{transformer_Vaswani2017}, our bi-modal encoder inputs two streams: audio $(A \in \R^{T_a \times d_a})$ and visual $(V \in \R^{T_v \times d_v})$ features corresponding to the proposal. Then, the features are passed in a stack of $N$ encoder layers. Instead of two, each layer has three sub-layers: \textit{self-attention}, \textit{bi-modal attention} (new), and \textit{position-wise fully-connected} layers. Specifically, given $A^\text{fc}_0 = A$ and $V^\text{fc}_0 = V$, an $n^{\text{th}}$ encoder layer is defined as
\begin{align}
    A^\text{self}_n &= \text{MultiHeadAttention}(A^\text{fc}_{n-1}, A^\text{fc}_{n-1}, A^\text{fc}_{n-1}), && \text{// audio self-attention}\\
    V^\text{self}_n &= \text{MultiHeadAttention}(V^\text{fc}_{n-1}, V^\text{fc}_{n-1}, V^\text{fc}_{n-1}), && \text{// visual self-attention}\\
    A^\text{mm}_n &= \text{MultiHeadAttention}(A^\text{self}_n, V^\text{self}_n, V^\text{self}_n), && \text{// visual-attended audio feats.} \label{eq:bimodal_att1}\\
    V^\text{mm}_n &= \text{MultiHeadAttention}(V^\text{self}_n, A^\text{self}_n, A^\text{self}_n), && \text{// audio-attended visual feats.} \label{eq:bimodal_att2}\\
    A^\text{fc}_n &= \text{TwoFullyConnected}(A^\text{mm}_n), && \text{// $ \R^{T_a \times d_a} \leftarrow \R^{T_a \times 4d_a} \leftarrow \R^{T_a \times d_a}$} \\
    V^\text{fc}_n &= \text{TwoFullyConnected}(V^\text{mm}_n), && \text{// $ \R^{T_v \times d_v} \leftarrow \R^{T_v \times 4d_v} \leftarrow \R^{T_v \times d_v}$}
\end{align}
where all sub-layers have distinct sets of trainable weights and mostly resemble the blocks of Transformer \cite{transformer_Vaswani2017}, yet we allow the dimension of the weights in multi-headed attention in \eqref{eq:bimodal_att1} \& \eqref{eq:bimodal_att2} to be different for both modalities because we expect them to have a different size. We define the multi-headed attention in Sec.~\ref{sec:building_blocks}. The encoder outputs visual-attended audio features $(A^v = A_N^\text{fc})$ and audio-attended visual features $(V^a = V_N^\text{fc})$, which are used\,the\,decoder.

\vspace{-1.4ex}\paragraph{Bi-modal Decoder \label{sec:decoder}}
The bi-modal decoder inputs the previous sequence of caption words $C_t = (c_1 \comma c_2 \comma \dots \comma c_t) \inR^{t \times d_c}$ and, opposed to the original Transformer's decoder \cite{transformer_Vaswani2017}, ours gets the output from the bi-modal encoder  $(A^v \inR^{T_a \times d_a} \comma V^a \inR^{T_v \times d_v})$. Thus, instead of three, it has four sub-layers: self-attention, \textit{bi-modal} encoder-decoder attention (new), \textit{bridge} (new), \& position-wise fully-connected layers. For $C^\text{fc}_0 = C_t$, an $n^\text{th}$ decoder layer is\,defined\,as
\begin{align}
    C^\text{self}_n &= \text{MultiHeadAttention}(C^\text{fc}_{n-1}, C^\text{fc}_{n-1}, C^\text{fc}_{n-1}), && \text{// caption self-attention} \\
    C^{A^v}_n &= \text{MultiHeadAttention}(C^\text{self}_n, A^v, A^v), && \text{// audio-visual attended prev. caps.} \\
    C^{V^a}_n &= \text{MultiHeadAttention}(C^\text{self}_n, V^a, V^a), && \text{// visual-audio attended prev. caps.} \\
    C^\text{mm}_n &= \text{OneFullyConnected}\big([C^{A^v}_n \comma C^{V^a}_n]\big), && \text{// $\R^{t \times d_c} \leftarrow \R^{t \times 2d_c}$; $[\cdot, \cdot]$ --- concat.} \\
    C^\text{fc}_n &= \text{TwoFullyConnected}(C^\text{mm}_n), && \text{// $ \R^{t \times d_c} \leftarrow \R^{t \times 4d_c} \leftarrow \R^{t \times d_c}$} \label{eq:decoder_out}
\end{align}
where, as in the encoder, trainable weights have distinct dimensions depending on a modality and are not shared across sub-layers. The decoder outputs caption features $(C^\text{av}_t=C^\text{fc}_N)$.

\vspace{-1.4ex}\paragraph{Generator} The purpose of the generator is to model the distribution for the next caption word $c_{t+1}$ given the output of the decoder $C^\text{av}_t \inR^{t \times d_c}$. Therefore, the generator is, usually, a fully-connected layer with the softmax activation which maps the caption features of size $d_c$ into a dimension corresponding to the size of the vocabulary in the training set.

\paragraph{Residual Connection}
Following the original Transformer architecture, we employ the \textit{residual connection} \cite{ResNet_He2016} surrounding each sub-layer of the encoder and decoder except for the bridge layer since in- and out-dimensions are different. Additionally, we adopt Layer Normalization \cite{LayerNorm_Ba2016} before applying a sub-layer: $x + \text{sub-layer}\big(\text{LayerNorm}(x)\big)$.

\vspace{-2.5ex}\paragraph{Dropout} We also regularize our model with \textit{dropout} \cite{dropout_Srivastava_2014} which is applied: a) before adding the residual in the residual connection, b) before the activation in the bridge layer, c) on outputs of the positional encoding, d) between layers in the position-wise fully-connected network, and e) after the softmax operation in the scaled dot-product attention (see Sec.~\ref{sec:building_blocks}).

\begin{figure}
\begin{minipage}[c]{0.60\textwidth}
    \includegraphics[width=\textwidth]{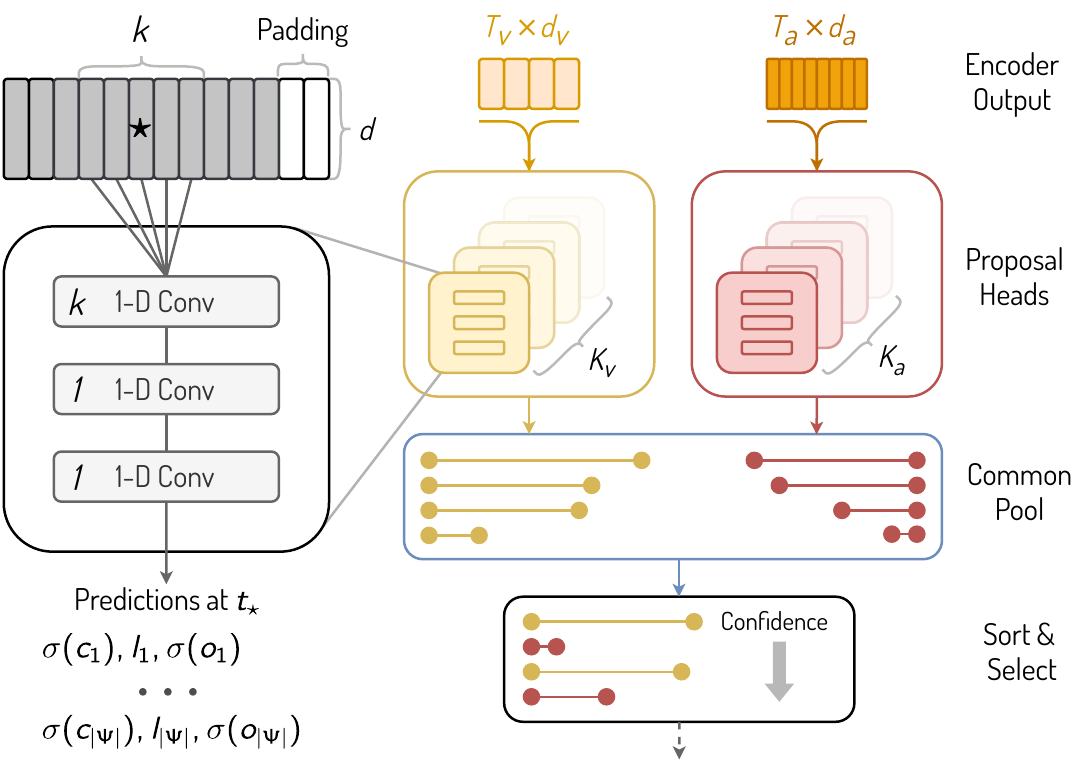}
\end{minipage}\hfill
\begin{minipage}[c]{0.38\textwidth}
\vspace{-3.5em}\caption{The Bi-modal Multi-headed Proposal Generator inputs the two-stream output from the bi-modal encoder, processes it with two stacks of proposal generation heads. The predictions from all heads form a common pool of predictions. Thus, the pool consists of $T_v \cdot K_v \cdot |\Psi_v| + T_a \cdot K_a \cdot |\Psi_a|$ proposals, which are sorted on the confidence score and passed back to clip input features to the captioning module. \label{fig:prop_gen}}
\vspace{-3em}
\end{minipage}
\end{figure}

\vspace{-1.3ex}\subsection{Event Proposal Generation Module\label{sec:prop_generator}}
\vspace{-0.2ex}The proposal generator generates a set of proposals for a given video. It consists of two blocks: a bi-modal encoder and \textit{bi-modal multi-headed proposal generator} (not related to multi-\textit{headed} attention). The bi-modal encoder in this module inputs the whole sequence opposed to the bi-modal encoder in the captioning module, which inputs a sequence of features corresponding to a proposal. Specifically, it inputs both: visual-attended audio features $A^v \inR^{T_a \times d_a}$ and audio-attended visual features $V^a\inR^{T_v \times d_v}$. Since the sequence lengths $(T_a \comma T_v)$ might be distinct, the fusion of predictions cannot be done at each time-step. To this end, we propose the module which makes predictions for each modality at every timestamp individually forming a common pool of cross-modal predictions (see Fig.~\ref{fig:prop_gen}).

\vspace{-2.3ex}\paragraph{Proposal Generation Head} The proposal generation head inputs a sequence of $T$ features, and makes predictions at each timestamp on the interval $[1\comma T]$, and for every prior segment length \textit{anchor} in the set $\Psi$. The design of the proposal generation head is partly inspired by \textit{YOLO} object detector \cite{YOLOv1_Redmon_2016,YOLOv2_Redmon_2016,YOLOv3_Redmon_2018}. Specifically, it is a \textit{fully-convolutional} network which, in our case, consists of only three layers. Opposed to YOLO, we preserve the sequence length across all layers using \textit{padding} and identity \textit{stride}. Moreover, YOLO utilizes predictions from three different scales to predict different-scale objects. Hence, only three sizes of receptive fields are used. Instead, our model makes predictions at a single scale while controlling the receptive field with a \textit{kernel size} $k$ which is distinct in each proposal generation head. More precisely, the $1^\text{st}$ convolutional layer has a kernel size $k$ while in the $2^\text{nd}$ and the $3^\text{rd}$ the kernel size is 1. The layers are separated with \textit{ReLU} activations and dropout.

\vspace{-2ex}\paragraph{Predictions} Temporal boundaries and confidence for a proposal are obtained using three values which were predicted by the proposal generation head: a location of a segment center $\sigma(c)$ relative to a position $p$ in the sequence while $\sigma(\cdot)$ is a sigmoid function which bounds the values into $[0 \comma 1]$ interval, a coefficient $\exp(l)$ for an anchor, and \textit{objectness score} $\sigma(o)$
\begin{align}
    \text{center} = p + \sigma(c); && \text{length} = \text{anchor} \cdot \exp{(l)}; && \text{confidence} = \sigma(o).\label{eq:preds}
\end{align}
The prediction of the center and length are in grid-cells (not in seconds). To obtain seconds, both are multiplied by a cell size which corresponds to a temporal span of the feature.

\vspace{-2ex}\paragraph{Bi-modal Multi-headed Proposal Generator}
The common pool of predictions is formed with predictions made by each of the proposal generation heads. Specifically, our model has $K_a$ and $K_v$ heads for audio and visual modalities with distinct sets of kernel sizes. Overall, our model generates $3 \cdot \big(T_a \cdot K_a \cdot |\Psi_a| + T_v \cdot K_v \cdot |\Psi_v|\big)$ proposals. For the final predictions, we select top-100 proposals out of the common pool based on the confidence score.

\vspace{-2ex}\paragraph{Segment Length Priors \& Kernel Sizes}
To select a set of anchors, we use \textit{K-Means} clustering algorithm with the \textit{Euclidean distance} metric, as opposed to \textit{intersection over the union} in YOLO. Due to granularity of feature extractors, feature lengths $(T_a \comma T_v)$ might not necessarily equal. Thus, we obtain distinct numbers of anchors for audio and visual modalities $\big(|\Psi_a| \comma |\Psi_v|\big)$ to keep $T_a \cdot |\Psi_a|$ close to $T_v \cdot |\Psi_v|$ to balance the impact of each modality to the common pool of predictions. Similarly, the kernel sizes are determined by K-Means. We motivate it with an expectation that the receptive field will correspond to an event with a higher probability. We scale the resulting cluster centroids (in secs) by the feature time span to obtain values in grid-cell coordinates. Next, we round the values to the next odd integer for more elegant padding. Again, to preserve the balance in the share of predictions from each modality, we obtain an equal number of kernel sizes $K_a = K_v$ both modalities.\vspace{-1ex}

\subsection{Training Procedure\label{sec:train_proc}}

\vspace{-1ex}Our model is trained in two stages: first, the captioning module is trained with ground truth proposals and, then, the proposal generator is trained using the pre-trained bi-modal encoder from the captioning model. Similar to \cite{transformer_Vaswani2017} and \cite{mdvc_Iashin_2020}, we optimize \textit{KL-divergence} loss and apply \textit{Label Smoothing} \cite{Szegedy2016} to force a model to be less confident about predictions anticipating noisy annotations. Also, \textit{masking} is used to ignore padding and prevent the model from attending to the next positions in the ground truth caption. During training of the event proposal generation module, all proposal generation heads for each modality are trained simultaneously summing up losses from all heads and both modalities. Each head uses YOLO-like loss: MSE for the localization losses (no square root) and \textit{cross-entropy} for (no)objectness losses. The NMS is avoided for efficiency and to preserve the possibility of \textit{dense} events. For the implementation details, a reader is referred to supplementary material (Sec.~\ref{sec:implementation}).

\section{Experiments}

\vspace{-1ex}We employ ActivityNet Captions dataset \cite{Krishna2017}, which consists of 100k temporally localized sentences for 20k YouTube videos. The dataset is split into 50/25/25\,\% parts for training, validation, and testing. The validation set of videos is annotated by two different annotators. We report the results on the validation subsets as ground truth is not available for the testing set. Since the dataset is distributed as a set of links to YouTube videos, it is not possible to collect the whole dataset as some videos became unavailable. The authors also provide C3D features which are not suitable for our experimentation as they are missing audio information. In total, we had 91\,\% of the videos. We omit the unavailable videos from the validation sets. We compared the results of other methods on the 91\,\% and 100\,\% of videos in Sec.~\ref{sec:other_models_no_missings} and observed similar performance suggesting the videos to be \textit{missing completely at random}.

To evaluate the event proposal generation module we employ precision, recall, and mainly rely on F1-score (harmonic mean of precision and recall). While METEOR \cite{METEOR_Denkowski2014} and BLEU@3--4 \cite{bleu_Papineni2002} were used for captioning as they are highly correlated with human judgement. All metrics are averaged for every video and \textit{temporal Intersection over Union} thresholds: $[0.3, 0.5, 0.7, 0.9]$. As it has been noted in \cite{Streamlined_Mun2019}, the original evaluation script had a critical issue which resulted in an incorrect evaluation of previous models. Therefore, we re-implement \cite{bafcg_Wang2018n,masked_transformer_Zhou2018} and compare with the results obtained with the corrected script.

\subsection{Comparison to the State-of-the-art}

\begin{table}
\small
\centering
\include{tables/dense_cap_results}
\caption{Comparison with state-of-the-art results on the dense video captioning task. The results are reported on the validation subset of ActivityNet Captions in both settings: captioning ground truth (GT) and learned proposals on BLEU@3--4 (B@3--4) and METEOR (M) metrics. For a fair comparison on METEOR, we additionally report the results of models without the reward (METEOR) maximization (RL) and indicate whether full dataset was available for training. The best and the 2$^\text{nd}$ best results are highlighted. \label{tab:dense_cap_results}}
\end{table}

We present the comparison between the bi-modal transformer with multi-headed proposal generator (Ours) and other methods in the existing literature \cite{Krishna2017,Li2018,masked_transformer_Zhou2018,bafcg_Wang2018n,MFT_Xiong2018,Streamlined_Mun2019,mdvc_Iashin_2020,Rahman2019} on the dense video captioning task. The results of the comparison for captioning both ground truth (GT) and learned proposals are shown in Tab.~\ref{tab:dense_cap_results}. Since evaluating captioning is still challenging and METEOR is probably the best among other options, yet it only provides a \textit{proxy} for how good a caption is. Therefore we believe that the direct optimization of METEOR using a reinforcement learning technique (RL) might not necessarily result in a better caption. To this end, we also include the results of \cite{Li2018,Streamlined_Mun2019} without the RL module. Moreover, we obtained the results of \cite{mdvc_Iashin_2020} on the same subset of videos as we have since they additionally removed the videos with no speech modality from the evaluation.

According to the results, in the learned proposals setup, our dense video captioning model outperforms all of the models, which have no reward maximization on METEOR (no RL) while being on par when captioning ground truth proposals. Notably, our model has the highest BLEU metrics in the learned proposal setup yet lies far away from \cite{masked_transformer_Zhou2018} when captioning ground truth proposals on BLEU and performs on par with this model\,on\,METEOR.

Comparing to the RL methods, our model still outperforms them on BLEU metrics in both setups but loses in METEOR due to the absence of reward-maximization module. We draw the attention of a reader to the performance of \cite{Li2018} with and without the RL module --- METEOR has dropped significantly yet other metrics remained on the same level.

Interestingly, we also outperform \cite{mdvc_Iashin_2020} who also use the transformer in multi-modal setup yet has more parameters (149M vs 51M). We note again that the results are not fair to neither of \cite{Rahman2019,mdvc_Iashin_2020} and ours since models have been trained on fewer videos.

\begin{table}
\begin{minipage}[c]{0.55\textwidth}
\centering
\small
\setlength\tabcolsep{0.5em}
\include{tables/proposal_gen_results}
\end{minipage}\hfill
\begin{minipage}[c]{0.38\textwidth}
\caption{Comparison with state-of-the-art proposal generation methods on dense video captioning task. Results are reported on the validation set of ActivityNet Captions. Metrics: Precision, Recall, \& F1-measure. The top-2 is highlighted. \label{tab:proposal_gen_results}}
\end{minipage}
\vspace{-3ex}
\end{table}

Next, we compare our bi-modal multi-headed proposal generation module with other proposal generation modules from other dense video captioning models. The results for \cite{masked_transformer_Zhou2018} and \cite{bafcg_Wang2018n} are reported for 100 proposals per video. The results of the comparison are presented in Tab.~\ref{tab:proposal_gen_results}. Despite our model being trained on fewer videos, our proposal generation model achieves state-of-the-art performance on the F1 metric. Specifically, our model provides impressive ground truth segment coverage while being accurate in its predictions.

\subsection{Ablation Study \label{sec:ablation_studies}}
In this section, we show how the training procedure and modality impact the final results. The results are presented in Tab.~\ref{tab:ablation_table} for both settings: captioning ground truth (performance of the captioning module) and leaned proposal (full dense video captioning model).\vspace{-2ex}

\paragraph{Training Procedures}
Our final model is trained in the following way. First, we train the captioning model on the ground truth proposal. Second, we freeze the weights on the encoder and train the proposal generator using the frozen encoder. The final results are obtained by captioning the proposals obtained from the trained proposal generator. Hence, the acronym ``Cap $\rightarrow$ Prop'' which reads as: ``the proposal generator is trained using the pre-trained encoder from the captioning module''. We compare this training procedure to other two methods: a) when both captioning and proposal generator modules are trained separately and b) when, first, the proposal module is trained and, then, the captioning module uses the pre-trained encoder with frozen weights during training. This is the opposite of the training procedure used for the final model, thus, abbreviated to ``Prop $\rightarrow$ Cap''.\vspace{-2ex}

\paragraph{Different Sets of Modalities}
The final model uses both audio and visual modalities to make predictions. We compare the performance of a bi-modal model with uni-modal ones. Specifically, for uni-modal settings, we employ the uni-modal transformer architecture similar to one in \cite{mdvc_Iashin_2020}. The difference between the hyper-parameters used for the final model and the uni-modal transformer is in the input dimension. For the uni-modal transformer, we follow the original paper where the input is first embedded into $D_q$ dimension (see \eqref{eq:attention_head}) and remains the same everywhere later. We select 1024 for visual-only and 128 for audio-only transformers; the size of the pre-trained GloVe is projected with a FC layer to match the size. \vspace{-4ex}

\paragraph{Results}
We report every combination of the settings in Tab.~\ref{tab:ablation_table}. Specifically, we observed that the captioning module does not benefit from the pre-training for the proposal generation (``Prop $\rightarrow$ Cap'' vs ``Cap $\rightarrow$ Prop'' \& ``Separate''). The results of the learned proposal setting show the importance of the pre-training but only in the ``Cap $\rightarrow$ Prop'' setting. Overall, we claim that the captioning training does not benefit from utilizing the pre-trained proposal generator's encoder and, even, performs worse with it. While, the proposal generator ends up with better performance if pre-trained captioning module's encoder is used.

The comparison of the cross-modal performance shows that using both modalities (audio and visual) gives the best result in nearly all cases in both settings. However, it is shown that the audio modality is the \textit{weakest} among the three implying that visual modality might contain a stronger signal for video understanding. Nevertheless, the gap between the visual-only and bi-modal case is consistent in all settings. This suggests that the audio still provides essential cues for dense video captioning. More ablations studies can be found in Sec.~\ref{sec:more_ablation_studies}.\vspace{-2ex}

\begin{table}
\centering
\include{tables/ablation}
\caption{The impact of training procedures and input modalities. We compare the training procedure of the final model when the proposal generator uses the pre-trained encoder on the captioning task (``Cap $\rightarrow$ Prop'') to an opposite scenario (``Prop $\rightarrow$ Cap''), and the situation when both of them are trained separately. The results are shown on validation sets of ActivityNet Captions when captioning ground truth (GT) and learned proposals. \label{tab:ablation_table}}
\vspace{-1.5ex}
\end{table}

\section{Conclusion}

\vspace{-1ex}We believe that the handling of multiple modalities is under-explored in the computer vision community. In this paper, we present a novel bi-modal transformer with a bi-modal multi-headed proposal generation module showing how audio might facilitate the performance of dense video captioning. We perform our experimentation on the ActivityNet Captions dataset and achieve state-of-the-art results on F1 and BLEU metrics. The the ablation study results show that the proposed model provides an effective and elegant way of fusing audio and visual features while outperforming the uni-modal configurations in all settings.\\[-1ex]

\noindent{\small\textbf{Acknowledgments}~\,Funding for this research was provided by the
Academy of Finland projects 327910 \& 324346. We also acknowledge CSC -- IT Center for Science, Finland, for computational~resources.}

\bibliography{src/bibliography}

\clearpage

\section{Supplementary Material}

\subsection{Multi-headed Attention \label{sec:building_blocks}}
\paragraph{Scaled dot-product attention} The notion of multi-headed attention is based on the idea of \textit{scaled dot-product attention} which is defined as follows
\begin{align}
    \text{Attention}(Q \comma K \comma V) = \text{Softmax}\bigg(\frac{Q K^{T}}{\sqrt{d}}\bigg) V,\label{eq:dot_product}
\end{align}
where $\sqrt{d}$ is a scaling factor designed to keep the Softmax gradients in a sufficient range, $Q \comma K \comma V$ are sequences of \textit{queries}, \textit{keys}, and \textit{values}, Softmax is applied row-wise.

\paragraph{Attention with Many Heads}
The concept of multiple heads was introduced in \cite{transformer_Vaswani2017} to allow a model to learn $H$ distinct representation sub-spaces at each position while preserving the same computation efficiency. An attention \textit{head} is usually presented as \eqref{eq:dot_product} with parametrized inputs
\begin{align}
    \text{head}_h(q, k, v) = \text{Attention}(qW^q_h \comma kW^k_h \comma vW^v_h), \quad h \in [1, H] \label{eq:attention_head}
\end{align}
where $q \inR^{T_q \times D_q} \comma k \inR^{T_k \times D_k} \comma v \inR^{T_k \times D_k}$ and $W^*_h \inR^{D_* \times D_\text{in}}$. Note that the inputs $k$ and $v$ are expected to have the same dimension $(T_k \times D_k)$ while $q$ might have a different one. The weights $W^*_h$ are mapping the corresponding inputs into an internal space $D_\text{in}=\frac{D_q}{H}$ such that $D_q$ is a multiple of $H$. The mapping into $D_\text{in}$ space allows the attention to be calculated between the features which originally were of distinct dimensions $(D_q \neq D_k)$. The multi-headed attention is, then, defined as the concatenation of $H$ attention heads mapped back to sub-space of queries $(D_q)$ with $W^\text{out} \inR^{H \cdot D_\text{in} \times D_q}$
\begin{align}
    \text{MultiHeadAttention}(q, k, v) = \big[\text{head}_1(q, k, v) \comma \text{head}_2(q, k, v) \comma \dots \comma \text{head}_H(q, k, v)\big] W^\text{out}.
\end{align}

\subsection{Feature Extraction \label{sec:feature_extraction}}

Both audio and visual features are pre-calculated before training. The audio features are extracted with the VGGish network \cite{Hershey2017}, which was pre-trained on \textit{AudioSet} \cite{Gemmeke2017}. More specifically, the VGGish model processes 0.96 seconds long segments. The audio segments, in turn, are represented as log mel-scaled spectrograms of size $96\times 64$ which are obtained via \textit{Short-time Fourier Transform}. The STFT utilizes a 25 ms \textit{Hann} window with 15 ms step applied to the 16 kHz mono audio track. The pre-classification layer of VGGish outputs a 128-d embedding for each spectrogram. Therefore, the audio track of an $i^\text{th}$ video in the dataset is represented with a sequence of 128-d features of length $T^i_a$, each feature in the stack represents 0.96 seconds of the original audio track.

To extract features from the visual stream, we employ the I3D network \cite{i3d_Carreira_2017} pre-trained on \textit{Kinetics} dataset. Specifically, I3D inputs 64 RGB and 64 optical flow frames of size $224^2$ extracted at 25 fps. Similar to \cite{mdvc_Iashin_2020}, we extract the flow frames using \textit{PWCNet} \cite{Sun2018PWC-Net}. Both sets of frames are, first, resized such that $\min(\text{Height}\comma \text{Width}) = 256$, and, then, the central region of size $224^2$ is cropped. After, both stacks of frames are passed through the corresponding streams of I3D. It outputs a 1024-d representation for RGB and flow 64-frame stacks from the second-to-the-last layer. Following the authors of I3D, we sum the representations from both streams. It results in a single 1024-d representation for every stack of 64 frames. Therefore, the visual track of $i\text{th}$ video is represented with a sequence of 1024-d features of length $T^i_v$ where every features spans 2.56 seconds (64 frames) of the original video.

The tokens (or, roughly, words) from captions are embedded with \textit{Global Vector} (GloVe) representations pre-trained on the \textit{Common Crawl} dataset (2.2M vocabulary) \cite{Glove_Pennington_2014}. The pre-trained model is represented as a lookup table which maps a token to a 300-d embedding. If a token is missing in the vocabulary, an average vector among all vocabulary words is returned. Therefore, each previous token of a caption is represented with a 300-d vector.

Therefore, the bi-modal encoder's in and out dimensions are $d_a=128$ and $d_v=1024$ for audio and visual streams while the decoder inputs and outputs $d_c=300$.

\subsection{Implementation Details \label{sec:implementation}}

The batch of size 32 and 16 were used during training of captioning and proposal generation modules, respectively. To form a batch, in the captioning module, the features are padded up to the longest sequence in the batch. For the proposal generator, the features are extracted from entire videos and padded up to 300 for visual and 800 for audio to form a batch. These number were selected to cover all possible lengths of the features in the training set. The padding is masked out as it is done for the next caption tokens in the decoder (see Sec.~\ref{sec:train_proc}). Each head in the bi-modal multi-headed generator predicts $\Psi_a=48$ and $\Psi_v =128 $ anchors for audio and visual modalities. We used the following lists each of size $K_a = K_v =10$ for kernel sizes (given in cell-coordinates): $[5, 13, 23, 35, 51, 69, 91, 121, 161, 211]$ for audio and $[1, 5, 9, 13, 19, 25, 35, 45, 61, 79]$ for visual modalities which are determined by K-Means algorithm. The size of both intermediate layers in proposal generation head is 512. Note that $\frac{128}{48}=\frac{800}{300}=\frac{2.56}{0.96}$ which preserves the balance between predictions from both modalities (see Sec.~\ref{sec:prop_generator}).

Since the modality features might have a different size, we also need to map them into an internal space inside of the bi-modal attention modules $(D_\text{in})$, see Eq.~\eqref{eq:attention_head} for more details. We select the internal space to be of size $D_\text{in}=1024$. Both the encoder and decoder of the bi-modal transformer have $N=2$ layers and $H=4$ heads in each of the multi-headed attention modules.  The caption vocabulary size and, hence, the generator's output dimension is 10\,172. We use $\gamma=0.7$ in the label smoothing and the probability of dropout $p=0.1$. The localization and objectness loss coefficients are 1, and the noobjectness coefficient is 100. \textit{Adam} optimizer with default hyper-parameters \cite{Kingma2014} and learning rate $5\cdot10^{-5}$ is used to train both caption and proposal generator. The hyper-parameters are selected on the validation\,set.

We highlight that the whole process of training both parts of the model was designed to keep a unified training procedure avoiding using different techniques such as \textit{reduce-on-plateau}, \textit{weight decay}, different learning rate, optimizer when training the proposal generator, sometimes, favoring elegance at the cost of performance. We encourage others to try different combinations when training both stages to achieve better results.

The captioning module was trained until METEOR on the validation set has not improved for 30 epochs while the proposal generator is trained for 70 epochs at most. In our experiments, the training of the final captioning module reaches the peak performance at 26$^\text{th}$ epoch while the proposal generator achieves the highest F1-score on the validation set at 17$^\text{th}$ epoch. We select the proposals on the epoch with the highest metric and caption them with the best captioning model. The training of the captioning module until the best performance takes 10 hours and 3.5 hours to train the proposal generator on \textit{one} Nvidia GeForce RTX 2080Ti. We use PyTorch \cite{PyTorch_Paszke_2019} as our primary library for the implementation.


\begin{table}
\centering
\include{tables/other_models_no_missings}
\caption{The performance of other methods on the filtered ActivityNet Captions validation set for videos which are no longer available (around 91\,\% (as ours)). The results are reported in the learned proposal setting. As expected, the performance of other models remains at the same level while ours gains the missing 9\,\%. Metrics are BLEU3--4, METEOR, recall, precision, and F1-measure. \label{tab:other_models_no_missings}}
\end{table}

\subsection{More Ablation Studies \label{sec:more_ablation_studies}}
\subsubsection{Why Do You Exclude Videos from the Validation Set?\label{sec:other_models_no_missings}}

In our experimentation, we exclude videos which are no longer available on YouTube (9\,\%) from the ground truth validation datasets as it would be unfair to compare our model to the methods which could make a prediction based on the video content while our model gets zero scores on a missing video. Therefore, we evaluate the predictions made by other models \cite{masked_transformer_Zhou2018,bafcg_Wang2018n} on the same validation set as we have. We selected only these two methods as they made either a code or evaluation results publicly available.

In other words, we hypothesise that the performance of other methods will not change after excluding videos from both predictions and ground truth while the performance of our method will be higher by around 9\,\% (a portion of the missing videos). The results of the comparison are shown in Tab.~\ref{tab:other_models_no_missings} and, indeed, imply that the performance of other methods remains on the same level (less than 2\,\% change). We remind a reader that the compared methods were trained on the full training dataset while ours was trained on only 91\,\%.

\begin{table}
\centering
\include{tables/how_our_model_improves_other_results}
\caption{The comparison of the captioning performance between our model and \cite{bafcg_Wang2018n} on the learned proposals provided in \cite{bafcg_Wang2018n}. The results are reported on the filtered ActivityNet Caption validation datasets. \label{tab:how_our_model_improves_other_results}}
\end{table}

\subsubsection{Might Your Model Improve Results of Other Methods?\label{sec:how_our_model_improves_other_results}}

Since \cite{bafcg_Wang2018n} have not made the results publicly available for captioning ground truth (see Tab.~\ref{tab:dense_cap_results}), we cannot compare it with our model directly. To this end, we apply our final captioning model on the generated proposals from \cite{bafcg_Wang2018n} to eliminate the effect caused by different proposal generator modules. The results of the comparison are reported on the filtered ActivityNet Caption validation datasets (see Sec.~\ref{sec:other_models_no_missings}) and shown in Tab.~\ref{tab:how_our_model_improves_other_results}. The results suggest that our model has a better captioning performance on this set of metrics.

\begin{figure}
\begin{center}
\includegraphics[width=\textwidth]{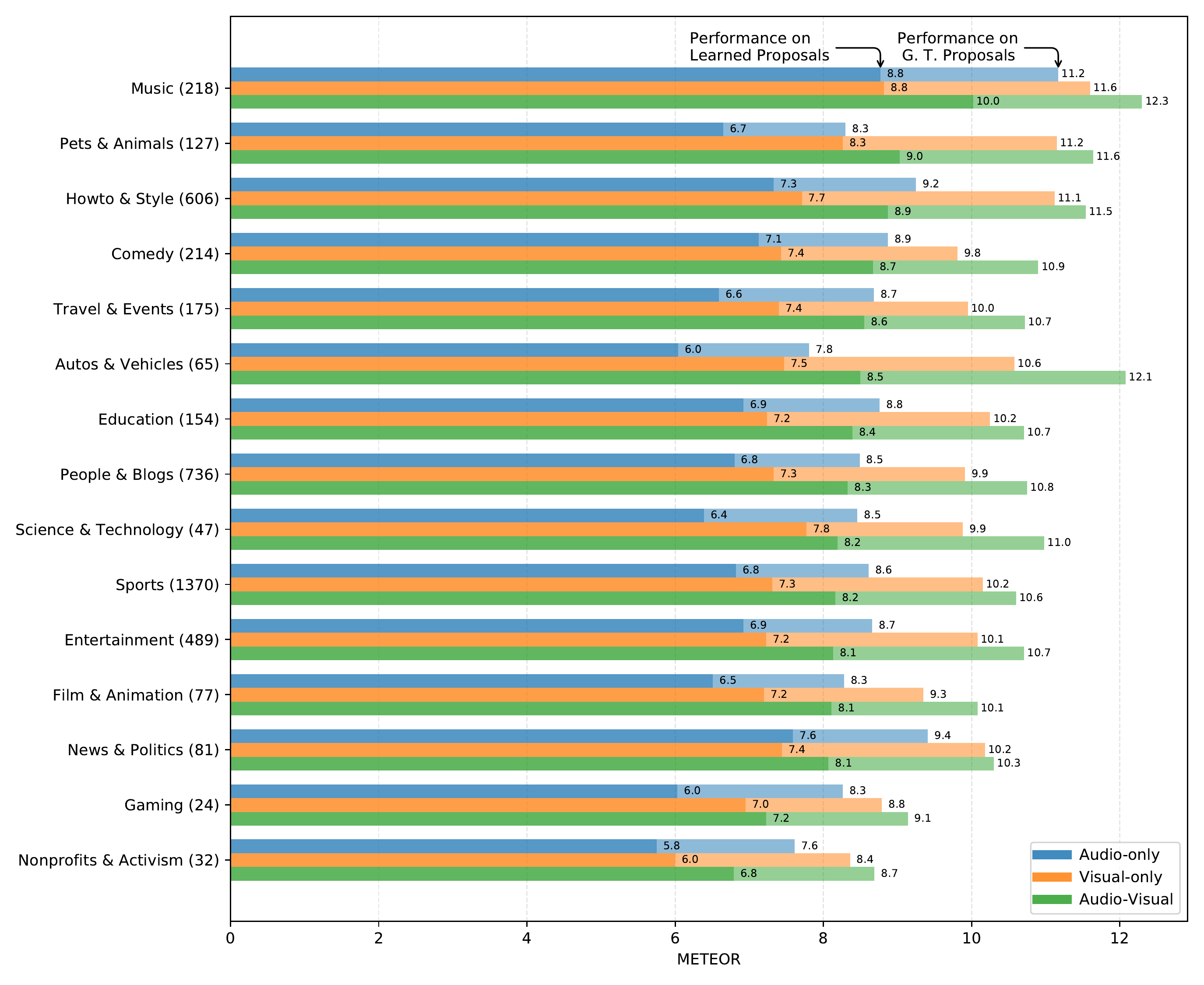}
\end{center}
\vspace{-1em}
\caption{The performance comparison between different modalities (Audio-only, Visual-only, and Bi-modal) in two settings (ground truth and learned proposals) across different YouTube video categories. The video categories are sorted according to the performance of the Audio-Visual model in the learned proposal setup. The number of videos in a category is shown in brackets. ActivityNet Captions validation subset is used for the comparison. \label{fig:performance_per_cat}}
\end{figure}

\subsubsection{What is the Impact of Audio and Visual Cues across Different Video Types?\label{sec:diff_cats}}

Following \cite{mdvc_Iashin_2020}, we inspect if the final model's performance consistently improves across different types of videos. To form a list of video types, we retrieve a YouTube video category for each video in the validation dataset. The YouTube category is annotated by the author when they upload a video to the service. YouTubeAPI \cite{YouTubeVideoCategories} was used to retrieve the categories automatically. Since there was a time gap between downloading videos and their categories, 67 were no longer available. Such videos were removed from the comparison. Also, we removed one video category with less than 20 videos.

Fig.~\ref{fig:performance_per_cat} shows the performance comparison between bi-modal (final), audio-only, and visual-only models across different video categories in two settings: captioning ground truth and learned proposals. The results suggest the consistent gain in performance when both modalities are used compared to the uni-modal models. This pattern holds across both settings and all categories. In addition, it appears that the visual modality provides more cues to the model than the audio modality in nearly all cases. Moreover, the dataset seems to be biased to ``Sports'' and ``People \& Blogs'' videos, which hold almost half of the dataset. Yet, the results show no evidence of over-fitting to these categories. Among all categories, ``Music'' appears to be the ``easiest'' one, which might be explained by a small variety of ways to describe the content of this kind. Meanwhile, the models perform the worst on ``Gaming'' and ``Nonprofits \& Activism'' categories, which might occur because of the lack of such videos in the dataset.

\begin{table}
    \begin{minipage}[c]{0.35\textwidth}
        \centering
        \setlength\tabcolsep{0.3em}
        \vspace{-1ex}\include{tables/no_bi_modal_att}
    \end{minipage}\hfill
    \begin{minipage}[c]{0.48\textwidth}
        \caption{The effect of replacing the bi-modal attention with a self-attention module in encoder layers. The comparison in shown on validation subsets of ActivityNet Captions in the learned proposal setting. The metrics are BLEU3--4 and METEOR. \label{tab:no_bi_modal_att}}
    \end{minipage}
\end{table}

\subsubsection{What Happens if the Bi-Modal Attention Block is Replaced by Uni-modal Self-Attention?\label{sec:no_bi_modal_att}}

In this ablation study, we would like to estimate the influence of the bi-modal attention blocks on the model performance (see middle blocks of Encoder and Decoder layers in Fig.~\ref{fig:bi_modal_transformer}). Yet, we can ablate only the encoder as the bi-modal attention is essential for the decoder since it inputs two streams. One solution would be to fuse the outputs of the encoder. This would, in turn, allow us to replace two bi-modal attention blocks in the decoder with one. However, it is not possible since the temporal spans of the encoder's output streams, in general, are distinct ($A^v\inR^{T_a\times d_a}$ and $V^a\inR^{T_v\times d_v}$). Therefore, in this setting, each encoder layer has two pairs of self-attention blocks which preserves the final model's number of parameters.

The results are presented in Tab.~\ref{tab:no_bi_modal_att}. We observed a substantial decline in performance among all metrics when the bi-modal attention block is replaced by self-attention. This further suggests the importance of the proposed approach. Besides, the results of the model with self-attention in encoder layers are still much stronger than any V- and A-only models (see Tab~\ref{tab:ablation_table}), which proves the importance of an audio-visual approach to the task.

\subsection{Qualitative Analysis \label{sec:qualitative_analysis}}

\begin{figure}[h]
\begin{center}
\includegraphics[width=\textwidth]{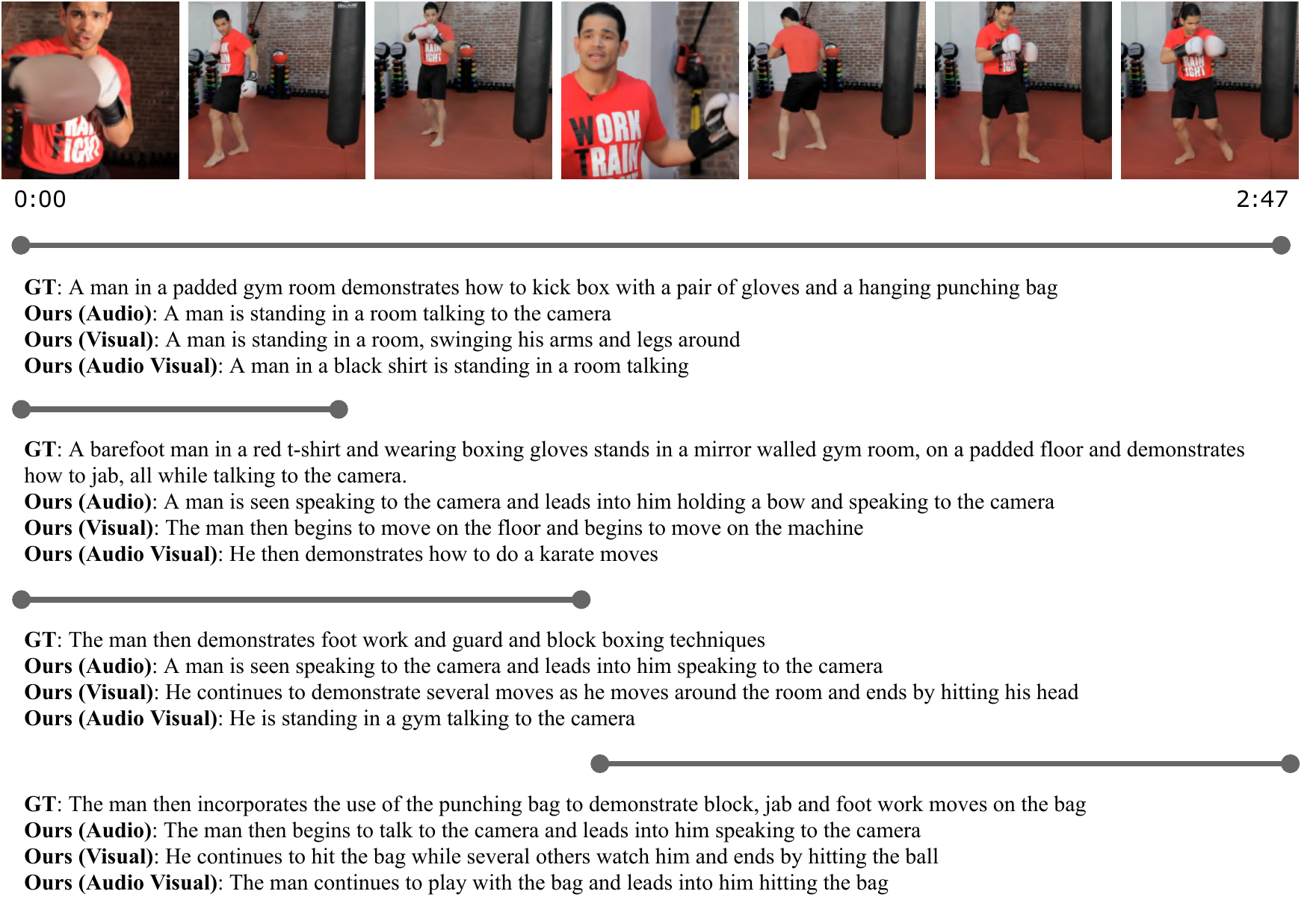}
\end{center}
\caption{The results of the qualitative analysis for a video from ActivityNet Caption validation dataset. The predictions of our bi-modal model are compared to the uni-modal model predictions and ground truth (GT) annotations. The video shows a man who explains how to do a martial art movement---the YouTube video id \texttt{EIibo7aTpys}. \label{fig:qualitative_analysis}}
\end{figure}

Fig.~\ref{fig:qualitative_analysis} provides the qualitative analysis of the final captioning model compared to ground truth captions. Additionally, we provide captions produced by uni-modal captioning models (audio- and visual-only). The results show that the caption, produced by a bi-modal captioning module as well as the audio-only model managed to grasp the concept of talking when captioning the largest segment (the top one) while video-only model neglects it. The video, by itself, consists of an explanation of how to do a martial art movement and highly verbose. Therefore, even though the ground truth does not mention that the person talks during the video, the predictions of our final model are not entirely erroneous. However, the colour of the man's shirt is incorrectly guessed, which might be explained by the presence of the black punching bag on the screen. Finally, the caption produced by the visual-only model also makes sense.

Moreover, if we consider the results of the audio-only model, we may notice that it mostly gets the signal of ``talking'' and exploits it in a prediction. Indeed, it might be challenging even for a non-English human to understand what the video is about given only the audio track. We also notice that captions provided by an annotator are significantly more detailed compared to the predictions of the captioning model, which are somewhat more general. This is the issue which needs more attention in future research as it seems to be a problem for any dense video captioning system.

\end{document}

%% file: tables/dense_cap_results.tex
\begin{tabular}{lcc rrr rrr}
\toprule
& & \textbf{Full Dataset} & \multicolumn{3}{c}{ \textbf{GT Proposals} } & \multicolumn{3}{c}{ \textbf{Learned Proposals} } \\
& \textbf{RL} & \textbf{was Available} & \textbf{B@3} & \textbf{B@4} & \textbf{M} & \textbf{B@3} & \textbf{B@4} & \textbf{M} \\
\midrule
Li \etal \cite{Li2018} & yes & yes & 4.55 & 1.62 & 10.33 & 2.27 & 0.73 & 6.93 \\
Xiong \etal \cite{MFT_Xiong2018} & yes & yes & -- & -- & -- & 2.84 & 1.24 & 7.08 \\
Mun \etal \cite{Streamlined_Mun2019} & yes & yes & 4.41 & 1.28 & \textbf{13.07} & \textbf{2.94} & 0.93 & \textbf{8.82} \\
\midrule
Krishna \etal \cite{Krishna2017} & no & yes & 4.09 & 1.60 & 8.88 & 1.90 & 0.71 & 5.69 \\
Li \etal \cite{Li2018} & no & yes & 4.51 & 1.71 & 9.31 & 2.05 & 0.74 & 6.14 \\
Zhou \etal \cite{masked_transformer_Zhou2018} & no & yes & \textbf{5.76} & \textbf{2.71} & \textbf{11.16} & 2.91 & \textbf{1.44} & 6.91 \\
Wang \etal \cite{bafcg_Wang2018n} & no & yes & -- & -- & 10.89 & 2.27 & 1.13 & 6.10 \\
Mun \etal \cite{Streamlined_Mun2019} & no & yes & -- & -- & -- & -- & -- & 6.92 \\
\midrule
Iashin \etal \cite{mdvc_Iashin_2020} & no & no & 4.52 & 1.98 & 11.07 & 2.53 & 1.01 & 7.46 \\
Rahman \etal \cite{Rahman2019} & no & no & 3.04 & 1.46 & 7.23 & 1.85 & 0.90 & 4.93 \\
Ours & no & no & \textbf{4.63} & \textbf{1.99} & 10.90 & \textbf{3.84} &\textbf{ 1.88} &\textbf{ 8.44} \\
\bottomrule
\vspace{0ex}
\end{tabular}

%% file: tables/proposal_gen_results.tex
\begin{tabular}{lc rrr}
\toprule
& \textbf{Full Dataset} & & & \\
& \textbf{was Available} & \textbf{Prec}. & \textbf{Rec}. & \textbf{F1} \\
\midrule
Xiong \etal \cite{MFT_Xiong2018} & yes & \textbf{51.41} & 24.31 & 33.01 \\
Wang \etal \cite{bafcg_Wang2018n} & yes  & 44.80 & 57.60 & 50.40 \\
Zhou \etal \cite{masked_transformer_Zhou2018} & yes & 38.57 & \textbf{86.33} & 53.31 \\
Mun \etal \cite{Streamlined_Mun2019} & yes & \textbf{57.57} & 55.58 & \textbf{56.56} \\
Ours & no & 48.23 & \textbf{80.31} & \textbf{60.27} \\
\bottomrule
\vspace{0ex}
\end{tabular}

%% file: tables/ablation.tex
\newcommand{\parlength}{5em}

\begin{tabular}{l l rrr rrr}
\toprule 
\textbf{Training} & & \multicolumn{3}{c}{ \textbf{GT Proposals} } & \multicolumn{3}{c}{ \textbf{Learned proposals} } \\
\textbf{Procedure} & \textbf{Modality} & \textbf{B@3} & \textbf{B@4} & \textbf{M} & \textbf{B@3} & \textbf{B@4} & \textbf{M} \\
\midrule 
\multirow{3}{*}{ \parbox{\parlength}{\small Separately} } & Audio & 2.85 & 1.14 & 8.81 & 2.50 & 1.11 & 6.89 \\
& Visual & 3.77 & 1.66 & 10.29 & 2.94 & 1.36 & 7.69 \\
& Bi-modal & 4.62 & 1.99 & 10.89 & 3.47 & 1.65 & 8.05 \\
\midrule
\multirow{3}{*}{ \parbox{\parlength}{\small Prop $\rightarrow$ Cap} } & Audio & 2.59 & 0.99 & 8.81 & 2.23 & 0.93 & 6.88 \\
& Visual & 3.62 & 1.56 & 10.16 & 3.08 & 1.45 & 7.81 \\
& Bi-modal & 4.10 & 1.78 & 10.48 & 3.07 & 1.47 & 7.67 \\
\midrule 
\multirow{3}{*}{ \parbox{\parlength}{\small Cap $\rightarrow$ Prop} } & Audio & 2.85 & 1.14 & 8.81 & 2.58 & 1.15 & 6.98 \\
& Visual & 3.77 & 1.66 & 10.29 & 2.85 & 1.30 & 7.47 \\
& Bi-modal & 4.62 & 1.99 & 10.89 & 3.84 & 1.88 & 8.44 \\
\bottomrule
\end{tabular}

%% file: tables/other_models_no_missings.tex
\newcommand{\parlength}{6.5em}

\begin{tabular}{lr rrr rrr}
\toprule 
& \textbf{Validation} & \textbf{B@3} & \textbf{B@4} & \textbf{M} & \textbf{Recall} & \textbf{Prec}. & \textbf{F1} \\
\midrule
\multirow{2}{*}{ \parbox{\parlength}{Wang \etal \cite{bafcg_Wang2018n}} } & Full & 2.27 & 1.13 & 6.10 & 57.60 & 44.80 & 50.40 \\
& As ours & 2.29 & 1.15 & 6.14 & 57.86 & 44.88 & 50.55 \\
\midrule
\multirow{2}{*}{ \parbox{\parlength}{Zhou \etal \cite{masked_transformer_Zhou2018}} } & Full & 2.91 & 1.44 & 6.91 & 86.33 & 38.57 & 53.31 \\
& As ours & 2.92 & 1.45 & 6.92 & 86.33 & 38.55 & 53.30 \\
\midrule
\multirow{2}{*}{ \parbox{\parlength}{Ours} } & Full & 3.50 & 1.72 & 7.69 & 73.22 & 43.97 & 54.95 \\
& As ours & 3.84 & 1.88 & 8.44 & 80.31 & 48.23 & 60.27 \\
\bottomrule
\end{tabular} 

%% file: tables/how_our_model_improves_other_results.tex
\begin{tabular}{l l rrr rrr}
\toprule
\textbf{Caption} & \textbf{Proposal} & & &  &  &  & \\
\textbf{Module} & \textbf{Generator} & \textbf{B@3} & \textbf{B@4} & \textbf{M} & \textbf{Recall} & \textbf{Prec.} & \textbf{F1} \\
\midrule
Wang \etal \cite{bafcg_Wang2018n} & \cite{bafcg_Wang2018n} & 2.29 & 1.15 & 6.14 & 57.86 & 44.88 & 50.55 \\
Ours & \cite{bafcg_Wang2018n} & 2.87 & 1.41 & 7.03 & 57.86 & 44.88 & 50.55 \\
Ours & Ours & 3.84 & 1.88 & 8.44 &  80.31 & 48.23 & 60.27 \\
\bottomrule
\end{tabular} 

%% file: tables/no_bi_modal_att.tex
\begin{tabular}{rccc}
\toprule
\textbf{2$^{\text{nd}}$ Encoder's Sub-layer} & \textbf{B3} & \textbf{B4} & \textbf{M} \\
\midrule
Self-Attention & 3.60 & 1.74 & 8.14 \\
Bi-modal Attention & 3.84 & 1.88 & 8.44\\
\bottomrule
\end{tabular}